\documentclass[letterpaper, 10 pt, conference]{ieeeconf}  

\IEEEoverridecommandlockouts                              

\usepackage{amsmath}
\usepackage{amssymb}
\usepackage{amsfonts}
\usepackage{url}
\usepackage[noend,ruled,vlined]{algorithm2e}
\usepackage{graphicx}
\usepackage{tikz}
\usetikzlibrary{shapes.geometric, arrows}
\usepackage{tabularx}
\usepackage{soul,xcolor}
\usepackage[normalem]{ulem}
\usepackage{units}

\usepackage[font=small,skip=2pt]{caption}

\DeclareMathOperator{\Tr}{Tr}
\DeclareMathOperator*{\argmin}{argmin}
\DeclareMathOperator*{\argmax}{argmax}

\newcommand{\subjectto}{\text{subject to}}
\usepackage{mathtools}
\DeclarePairedDelimiter\norm{\lVert}{\rVert}

\newcommand{\naturals}{\mathbb{N}}

\newcommand{\reals}{\mathbb{R}}
\newcommand{\se}[1]{SE(#1)}
\newcommand{\tanse}[1]{\mathfrak{se}(#1)}

\newcommand{\eye}{\mathbf{I}}

\newcommand{\ones}{\mathbf{1}}
\newcommand{\prob}{\mathsf{P}}

\newcommand{\name}[0]{CDCPD2}

\newcommand{\ngrippers}{G}
\newcommand{\gconfig}{q}
\newcommand{\gvel}{\dot{\gconfig}}

\newcommand{\obstacles}{\mathcal{O}}

\newcommand{\ndeform}{M}
\newcommand{\gmmidx}{m}
\newcommand{\Deformpoints}{\mathcal{P}}
\newcommand{\deformpoint}{p}
\newcommand{\Deformedges}{\mathcal{E}}
\newcommand{\deformedge}{e}
\newcommand{\geodistij}{\rho_{ij}}
\newcommand{\Deformprev}{\Deformpoints^{\textrm{prev}}}
\newcommand{\deformprev}{\deformpoint^{\textrm{prev}}}
\newcommand{\Deformgmm}{\Deformpoints^{\text{GMM}}}
\newcommand{\deformgmm}{\deformpoint^{\text{GMM}}}
\newcommand{\Deformpred}{\Deformpoints^{\textrm{pred}}}
\newcommand{\deformpred}{\deformpoint^{\textrm{pred}}}
\newcommand{\deformnear}[1]{\deformpoint_{\textrm{near},#1}}
\newcommand{\deformprevnear}[1]{\deformpoint^{\textrm{prev}}_{\textrm{near},#1}}

\newcommand{\nmasked}{N}
\newcommand{\maskedidx}{n}
\newcommand{\Maskedpoints}{\mathcal{D}}
\newcommand{\maskedpoint}{d}
\newcommand{\correspondences}{\mathcal{C}}
\newcommand{\correspondidx}{\mathcal{C}_{\textrm{idx}}}

\newcommand{\noiseweight}{w}
\newcommand{\posteriorprob}{X}
\newcommand{\lle}{L}
\newcommand{\kernel}{G}
\newcommand{\cpdw}{W}
\newcommand{\distnm}{\norm*{\maskedpoint_\maskedidx - \deformpoint_\gmmidx}}
\newcommand{\expdist}{\exp\left( -\frac{\distnm^2}{2 \sigma^2} \right)}

\newcommand{\gaussiannormalizerdenom}{(2\pi\sigma^2)^{\frac{3}{2}}}
\newcommand{\gaussiannormalizer}{\frac{1}{\gaussiannormalizerdenom}}

\newcommand{\prodovermasked}{\prod_{\maskedidx=1}^{\nmasked}}
\newcommand{\sumoverdeform}{\sum_{\gmmidx = 1}^{\ndeform}}
\newcommand{\sumovergmm}{\sum_{\gmmidx = 1}^{\ndeform + 1}}
\newcommand{\sumovermasked}{\sum_{\maskedidx = 1}^{\nmasked}}
\newcommand{\sumovermaskeddeform}{\sum_{\maskedidx, \gmmidx = 1}^{\nmasked, \ndeform}}
\newcommand{\maskedprobindiv}{\prob(\maskedpoint_\maskedidx)}
\newcommand{\maskedprobindivcond}{\prob(\maskedpoint_\maskedidx | \gmmidx)}
\newcommand{\maskedprob}{\prodovermasked \maskedprobindiv}
\newcommand{\maskedprobfull}{\prodovermasked \sumovergmm \prob(\gmmidx) \maskedprobindivcond}
\newcommand{\probgmmgivenmasked}{\prob(\gmmidx | \maskedpoint_\maskedidx)}
\newcommand{\prevsigma}{\sigma_{\text{prev}}^2}
\newcommand{\diffsigma}{\sigma_{\text{diff}}^2}

\newcommand{\yixuan}{Yixuan~Wang}
\newcommand{\dale}{Dale~M\textsuperscript{c}Conachie}
\newcommand{\dmitry}{Dmitry~Berenson}

\title{\LARGE \bf%
Tracking Partially-Occluded Deformable Objects while Enforcing Geometric Constraints}

\author{\yixuan$^{1}$, \dale$^{1}$, and \dmitry$^{1}$
\thanks{$^{1}$\yixuan, \dale, and \dmitry\ are with the University of Michigan, Ann Arbor, MI, USA.
        {\tt\small \{yixuanwa, dmcconac, dmitryb\}@umich.edu}}%
}

\begin{document}

\maketitle
\thispagestyle{empty}
\pagestyle{empty}

\begin{abstract}
In order to manipulate a deformable object, such as rope or cloth, in unstructured environments, robots need a way to estimate its current shape. However, tracking the shape of a deformable object can be challenging because of the object's high flexibility, (self-)occlusion, and interaction with obstacles. Building a high-fidelity physics simulation to aid in tracking is difficult for novel environments. Instead we focus on tracking the object based on RGBD images and geometric motion estimates and obstacles. Our key contributions over previous work in this vein are: 1) A better way to handle severe occlusion by using a motion model to regularize the tracking estimate; and 2) The formulation of \textit{convex} geometric constraints, which allow us to prevent self-intersection and penetration into known obstacles via a post-processing step. These contributions allow us to outperform previous methods by a large margin in terms of accuracy in scenarios with severe occlusion and obstacles.


\end{abstract}

\vspace{-0.05in} 
\section{Introduction}
\vspace{-0.05in} 

Tracking the shape of a deformable object is a long-standing problem that has been studied for applications in computer graphics \cite{graphics:White, graphics:Li}, surgery \cite{surgical:Collins, surgical:heterogeneous}, computer vision \cite{cv:DynamicFusion, cv:featureBase}, and robotics \cite{robot:CDCPD, robot:TangTrack, robot:TangFramework}. However, tracking deformable objects in the presence of severe occlusion and obstacles remains a difficult open problem. The key challenge is to maintain an estimate of the shape that conforms with physical constraints, i.e. that the object's motion conforms to a reasonable motion model and that the object cannot move through obstacles or through itself. Occlusion makes enforcing these kinds of constraints especially difficult.

Previous work on this problem has explored using a physics simulator to inform the prediction estimate \cite{robot:TangState}, but such methods assume a simulation environment for the given scene (with appropriate friction, stiffness, etc. parameters) can be easily constructed. Even if this were the case, such methods are quite sensitive to occlusion. More recently, CDCPD \cite{robot:CDCPD} showed that such models were not necessary for accurate tracking, instead using geometric methods to infer the shape of occluded parts of the object. However, CDCPD has two key limitations: 1) It has no motion model to constrain how the object moves between frames and, as a result, large erroneous changes in state are possible due to occlusion; and 2) There is no way to enforce geometric constraints for self-intersection and obstacle penetration.

This paper builds on CDCPD to address these key deficiencies. Specifically, our method, \name{}, makes three novel additions: 1) We incorporate a user-defined motion model into the regularization process to bias the tracker toward realistic object motion; 2) We formulate convex constraints that prevent edge-crossing and incorporate them into a post-processing step; and 3) We formulate convex obstacle-penetration constraints and include them in post-processing. While we use a similar expectation-maximization procedure as CDCPD, we emphasize that the above additions are highly non-trivial and, more importantly, that they allow us to track deformable objects in much more realistic scenarios, where occlusion and clutter are unavoidable.

Our experiments compare our method to CDCPD and \cite{robot:TangState}. Our simulation results (where we can obtain the ground-truth state of the object) show that \name{}  is able to estimate the state of the object much better than the previous methods under severe occlusion. Our real-world results show a clear qualitative improvement in the estimate of the deformable object shape in several challenging scenarios. Our code is available open-source\footnote[2]{ \url{https://github.com/UM-ARM-Lab/cdcpd/tree/CDCPD2}}.


\begin{figure}[t]
    \centering
    \includegraphics[width=\columnwidth]{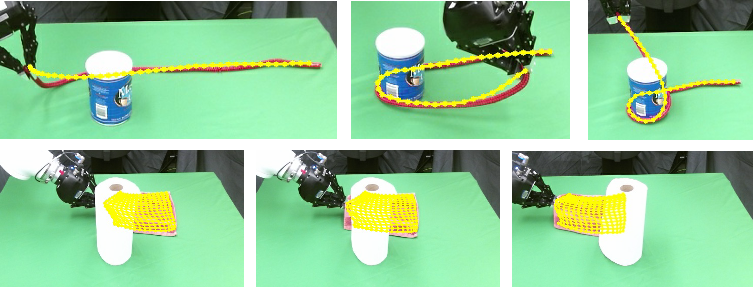}
    \caption{Examples of tracking results for rope winding around a cylinder (top) and cloth behind an obstacle (bottom). The 3D tracking estimate (yellow) is overlaid on the corresponding image.}
    \label{fig:intro}
\end{figure}

\vspace{-0.05in} 
\section{Related Work}
\vspace{-0.05in} 


Deformable object tracking has been studied in a variety of fields. Due to space constraints, we focus on only the most relevant methods to our work. In computer vision, \cite{cv:DynamicFusion} explores methods to reconstruct non-rigid scenes in real-time. However, it does not track a soft object model explicitly, as new nodes can be added and removed. 
For self-occlusion, \cite{cv:featureBase} introduces a method to handle self-occlusion based on feature matching between between a template and input images. However, the deformable objects we consider don't have obvious features to use for matching.

Deformable object tracking has also been used for image-guided surgery. However, these approaches typically rely on textural features \cite{surgical:Collins} or a high-fidelity simulation of the deformable object to inform tracking \cite{surgical:heterogeneous}. In contrast, our method does not require either of these and thus it is applicable to a broader range of objects and novel scenarios.

In robotics, \cite{robot:texture} introduces the method of representing a textured wire as a NURBS spline model and manipulates the wire based on this model. 
However, it requires the wires to have a specific color pattern. \cite{robot:realTime} focuses on tracking with obstacles and occlusion. However, it is based on the known Finite Element Method (FEM) model of the specific object, which can be difficult to construct. \cite{robot:clothGrasp} focuses on grasping cloth, where the tracking of the cloth is based on corner detection. \cite{robot:Knot} tracks and manipulates rope but is specialized to knot tying applications, whereas we seek to track both rope and cloth-like objects. \cite{robot:un-knotting, robot:Tangled} follows a similar approach, using hand-engineered features for perception.

Unlike much of the above work, which uses hand-engineered features for perception, several recent methods use point registration based on Gaussian Mixture Model Expectation-Maximization (GMM-EM) and Coherent Point Drift (CPD) (e.g. \cite{robot:trackPC}). Tang \textit{et al.} proposes to incorporate physics simulation into the point registration process \cite{robot:TangTrack, robot:TangFramework, robot:TangState}. This method is effective for knotting and folding, but it performs poorly when the object is significantly occluded (by itself or by obstacles) and it requires setting up a simulation with many friction parameters and material properties that are difficult to determine for a given deformable object. Chi and Berenson \cite{robot:CDCPD} introduce a method to track rope and cloth while allowing for occlusions of the object without relying on a physics simulator. However, their method does not address obstacle interactions and self-intersection. We compare to both \cite{robot:CDCPD} and \cite{robot:TangState} in our experiments.

\vspace{-0.05in} 
\section{Problem Statement}
\vspace{-0.05in} 

Let $\langle \Deformpoints^t, \Deformedges \rangle$ to be the configuration of the deformable object at time $t$. $\Deformpoints^t \in \reals^{3\ndeform}$ is a set of $\ndeform$ points, and every element $\deformpoint_i \in \Deformpoints^t$ is a point with coordinate $(x, y, z)$. $\Deformedges \in \mathbb{I}^{2E}$ is a set of $E$ edges, and every element $\deformedge_i \in \Deformedges$ has two indices of $\Deformpoints^t$ which form an edge. We assume that we know the poses of the robot's  $\ngrippers$ gripper(s). The gripper(s) configuration at time $t$ is denoted as $\gconfig^t \in \se3^G$ with velocity $\dot{\gconfig}^t$. Each element $\gvel_g \in \gvel^t$ can be written as $[v_g^T\ \omega_g^T]^T \in \tanse3$, i.e. the concatenation of the translational and angular velocities. We denote indices of $\Deformpoints^t$ grasped by $\ngrippers$ grippers at time $t$ as $\correspondences^t \in \mathbb{I}^{d\ngrippers}$, where $c_i\in \correspondences^t$ means $d$ indices grasped by the $i$th gripper. In addition, we define $\geodistij$ as the geodesic distance distance between points $\deformpoint_i$ and $\deformpoint_j$ on the surface of the deformable object.

The inputs to the tracking algorithm are a sequence of RGBD images $\mathcal{I}^t$, $\gconfig^t$ (optional), $\dot{\gconfig}^t$ (optional), $\correspondences^t$ (optional), and an initial connectivity model $\langle \Deformpoints^0, \Deformedges \rangle$. The optional may be used by some motion models, but it depends on the model. As with previous work (\cite{robot:TangState,robot:CDCPD}), we assume a point-cloud is generated from the RGBD image and the object is segmented out of the pointcloud. In this work, all points not belonging to the deformable object are assumed to be obstacles.
We denote deformable object masks as $\mathcal{M}^t$, with the same size as RGBD images $\mathcal{I}^t$, and masked deformable objects' corresponding point clouds are denoted as $\Maskedpoints^t\subset \reals^{3\nmasked}$, a set of $\nmasked$ points, and every element $\maskedpoint_i\in \Maskedpoints^t$ is a point with coordinate $(x, y, z)$. We assume that the obstacle points can be used to infer the shape of the obstacles (e.g. using shape-completion \cite{cv:shapeComp1, cv:shapeComp2}, model registration \cite{cv:icp}, or simply creating a mesh from the points), and thus we obtain a mesh representation of obstacles $\obstacles$.

Tracking a deformable object can then be regarded as a point registration problem of estimating $\Deformpoints^{t+1}$ by aligning $\Deformpoints^t$ to $\Maskedpoints^{t+1}$, with known obstacles $\obstacles$ and (optionally) gripper information $\langle \gconfig^{t+1}, \gvel^{t+1}, \correspondences^{t+1} \rangle$. Denoting the true configuration of the deformable object at time $t$ as $\Deformpoints^{*t+1}$, our goal is to output the $\Deformpoints^{t+1}$ that minimize $\norm{\Deformpoints^{*t+1} - \Deformpoints^{t+1}}_2$. In addition, we wish to obtain an estimate with the same topology (in the sense of the over- and under- crossings considered in knot theory \cite{robot:Knot}) as the ground truth. This problem is difficult because the tracking method must compensate for (self-)occlusion, prevent self-intersection, and enforce the geometric constraints arising from obstacles.


\vspace{-0.05in} 
\section{Method}
\vspace{-0.05in} 



As a foundation for our method, we build on Constrained Deformable Coherent Point Drift (CDCPD)~\cite{robot:CDCPD}. At its core, CDCPD uses a Gaussian Mixture Model (GMM) Expectation-Maximization (EM) process to find the configuration of the deformable object $\Deformpoints^{t+1}$ that best explains the observation $\Maskedpoints^{t+1}$ given the previous configuration $\Deformpoints^t$ and the robot configuration $\gconfig^{t+1}$ (Sec.~\ref{sec:gmmem}). In addition, CDCPD uses Coherent Point Drift (Sec.~\ref{sec:cpd}) and Locally Linear Embedding (Sec.~\ref{sec:lle}) to encourage the result to be physically plausible, however this is insufficient for some types of severe occlusion. In this paper, our first contribution is to introduce an additional novel regularization term based on a geometric model-based prediction (Sec.~\ref{sec:pred}); this enables us to use the motion of the robot $(\gconfig^t \rightarrow \gconfig^{t+1})$ to help infer the motion of the deformable object $(\Deformpoints^t \rightarrow \Deformpoints^{t+1})$.

While these regularization terms encourage physically-plausible results, they do not exploit constraints that we know about the system. CDCPD enforces constraints based on the position of the robot grippers and the maximum size of the deformable object (Sec.~\ref{sec:stretch}). However, CDCPD does not account for self-intersection and obstacles. Thus our second contribution is to introduce two additional constraints: a self-intersection constraint (Sec.~\ref{sec:self_inter}) that ensures that the tracked deformable object does not pass through itself, and an obstacle interaction constraint (Sec.~\ref{sec:obstacle}) that ensures that the tacked object is never inside an obstacle. The overall method is show in Alg. \ref{alg:method}.

\begin{algorithm}[t]
    \SetAlgoLined
    \caption{\name{}}
    \SetKwInOut{Input}{Input}
    \SetKwInOut{Output}{Output}

    \Input{$\Deformpoints^0, \Deformedges, \obstacles, \correspondidx^t, \mathcal{I}^t, \mathcal{M}^t, \gconfig^t, \gvel^t, \correspondences^t, t=1,2,...$}
    \Output{$\Deformpoints^t, t=1,2,...$}
    
    Compute $\kernel$ using Eq. \ref{eq:kernel}\;
    Compute $\lle$ using \cite{ml:LLE}\;
    \For{$t=0,1,...$}{
        Retrieve the masked point cloud $\Maskedpoints^{t+1}$ from $\mathcal{I}^{t+1}$\;
        $\Deformpoints^{\textrm{GMM}, t+1} = \text{GMM-EM}(\Deformpoints^t, \Maskedpoints^{t+1}, \lle, \kernel, \gconfig^{t+1}, \gvel^{t+1}, \correspondences^{t+1}, \obstacles)$\;
        Optimize $\Deformpoints^{t+1}$ using Eq. \ref{eq:constrain}\;
    }
    \label{alg:method}
\end{algorithm}

\vspace{-0.05in} 
\subsection{Deformable Object Tracking as Point Set Registration}
\vspace{-0.05in} 
\label{sec:gmmem}

Assuming relatively small changes in object shape between two consecutive frames, it is reasonable to regard the tracking problem as a point registration problem. In other words, we will align $\Deformpoints^t$ to $\Maskedpoints^{t+1}$ to get $\Deformpoints^{t+1}$. For simplicity, in this section, we will write $\Deformpoints^t$ as $\Deformprev$, $\Deformpoints^{t+1}$ as $\Deformpoints$, $\Deformpoints^{\textrm{GMM}, t+1}$ as $\Deformgmm$, and $\Maskedpoints^{t+1}$ as $\Maskedpoints$. Following the formulation in \cite{ml:CPD,cv:GLTP,robot:CDCPD}, the problem will initially be formulated as a naive GMM problem.

We assume every point $\deformpoint_i \in \Deformgmm$ is the center of a Gaussian distribution with the same isotropic variance $\sigma^2$. The sum of $\ndeform$ Gaussians is a Gaussian Mixture Model (GMM). In addition to the Gaussian terms, we add a uniform term with weight $\noiseweight \in (0, 1)$ to the mixture. Thus
\begin{equation}
    \prob(\gmmidx) = 
    \begin{cases}
        \frac{1-\noiseweight}{\ndeform},    & \gmmidx = 1, \dots, \ndeform\\
        \noiseweight,                       & \gmmidx = \ndeform + 1
    \end{cases}
    \label{eq:gmmprior}
\end{equation}
and
\begin{equation}
    \maskedprobindivcond  = 
    \begin{cases}
        \gaussiannormalizer \expdist,   & \gmmidx = 1, \dots, \ndeform\\
        \frac{1}{\nmasked},             & \gmmidx = \ndeform + 1
    \end{cases}
\end{equation}
We then assume that points in $\Maskedpoints$ were generated by drawing samples from the mixture:
\begin{align} 
    \label{likelihood}
    \prob(\Maskedpoints) = \maskedprob = \maskedprobfull \enspace .
\end{align}

Then our goal is to find the configuration of the deformable object that best explains the observation; i.e.
\begin{equation}
    \argmax_{\Deformpoints} \log \prob(\Maskedpoints) \enspace .
\end{equation}
 The EM algorithm can be used to solve this problem, as shown in~\cite{ml:CPD}. In the E-step we keep the Gaussian distributions fixed and calculate the expectation that $\maskedpoint_\maskedidx$ is generated by the $\gmmidx$'th Gaussian distribution
\begin{align}
    \probgmmgivenmasked =
    \begin{cases}
        \frac{1}{\eta} \expdist,                                                                                & \gmmidx = 1, \dots, \ndeform \\
        \frac{1}{\eta} \frac{\noiseweight}{(1-\noiseweight)}\frac{\gaussiannormalizerdenom \ndeform}{\nmasked}, & \gmmidx = \ndeform + 1
    \end{cases}
\end{align}
where $\eta > 0$ is set so that $\sumovergmm \probgmmgivenmasked = 1$. Then in the M-step, following the algorithm described in~\cite{ml:CPD,cv:GLTP}, we update the position of the Gaussian centroids $\Deformpoints$ and variance $\sigma$ by minimizing 
\begin{align}
    Q(\Deformpoints, \sigma^2) = \sumovermasked \sumovergmm \probgmmgivenmasked \frac{\distnm^2}{2\sigma^2} + \frac{3}{2}N_{\prob} \log \sigma^2
\end{align}
where $N_{\prob}$ is the sum of $\probgmmgivenmasked$, i.e. $\sum_{\maskedidx,\gmmidx=1}^{\nmasked, \ndeform} \probgmmgivenmasked$. Then we iterate E-step and M-step until convergence.

\begin{algorithm}[t]
    \SetAlgoLined
    \caption{GMM-EM($\Deformprev, \Maskedpoints, \lle, \kernel, \gconfig, \gvel, \correspondences, \obstacles$)}
    \SetKwInOut{Input}{Input}
    \SetKwInOut{Output}{Output}
    
    \Input{$\Deformprev, \Maskedpoints, \lle, \kernel, \gconfig, \gvel, \correspondences, \obstacles$}
    \Output{$\Deformgmm$}
    
    Compute $p_{vis}$ using Eq. (7) in \cite{robot:CDCPD}\;
    $\prevsigma=0$\;
    $\sigma^2=\text{Var}(\Deformprev)$\;
    $\diffsigma=\epsilon+1$\;
    $\cpdw = 0$\;
    $i=0$\;
    \While{$\diffsigma > \epsilon$ and $i<\text{max\_iter}$}{
        Compute $\posteriorprob$ using Eq. (9) in \cite{robot:CDCPD}\;
        Compute $\Deformpred$ using motion model\;
        Solve $W$ using Eq. \ref{eq:WSolver}\;
        $\prevsigma = \sigma^2$\;
        Compute $\sigma^2$ using Eq. \ref{eq:sigmaSolver}\;
        $\diffsigma=\text{abs}(\sigma^2 - \prevsigma)$\;
        $i++$\;
    }
    \Return{$\Deformgmm = \Deformprev + \kernel\cpdw$}
    \label{alg:gmmem}
\end{algorithm}

\subsubsection{Visual Information Exploited for Occlusion}
\label{sec:visprior}

The base GMM-EM formulation assumes a uniform prior for each Gaussian centroid (see Eq.~\ref{eq:gmmprior}); CDCPD replaces this assumption based on visibility information. If a particular point on the deformable object is occluded, then that point is unlikely to generate any samples in $\Maskedpoints$. This information is encoded in the posterior probability matrix $\posteriorprob \in \reals^{\ndeform \times \nmasked}$ (see Eq.~(9) in~\cite{robot:CDCPD}).

\vspace{-0.05in} 
\subsection{GMM-EM Regularization}
\vspace{-0.05in} 
\label{sec:reg}

\subsubsection{Coherent Point Drift Regularization}
\label{sec:cpd}

While GMM-EM forms a reasonable basis for a probabilistic interpretation of an observation; it does not account for dependencies between each Gaussian centroid. In our domain, there is physical structure to these centroids based on the deformable object; in particular two points on the deformable object that have a short geodesic distance between them are likely to move coherently. CPD encodes this structure by restricting the motion of the deformable object to be of the form
\begin{equation}
    \Deformgmm = \Deformprev + \kernel \cpdw
    \label{eq:cpd_transform}
\end{equation}
where $\kernel$ is a Gaussian kernel matrix with elements
\begin{equation}
    \kernel_{ij} = \exp \left( -\nicefrac{\geodistij^2}{(2\beta^2)} \right)
    \label{eq:kernel}
\end{equation}
and $\cpdw \in \reals^{\ndeform \times 3}$ is a weight matrix. We can then regularize this weight matrix to enforce motion coherence during the the M-step:
\begin{equation}
    E_{CPD} = \Tr \left( \cpdw^T \kernel \cpdw \right) \enspace .
    \label{eq:cpd_cost}
\end{equation}
Note that we deviate from CDCPD by using a static value for $\kernel$ rather than recalculating it at every timestep.

\subsubsection{Locally Linear Embedding Regularization}
\label{sec:lle}
While CPD is able to encourage motion coherence during consecutive frames, it does not account for excessive change over time. For example in the extreme case, two points can drift arbitrarily far apart given enough time. To address this problem CDCPD uses a regularization term proposed in~\cite{cv:GLTP}. This term is based on Locally Linear Embedding (LLE)~\cite{ml:LLE}, which represents each point as a linear combination of its $k$ nearest neighbours, and then penalizes deviation from that representation. We obtain linear weights $\lle$ by minimizing the following cost function:
\begin{equation}
    J(\lle) = \sumoverdeform \norm{\deformpoint_\gmmidx^{0} - \sum_{i \in K_\gmmidx} \lle_{\gmmidx i}\deformpoint_i^{0}}^2
\end{equation}
where $K_\gmmidx$ is a set of indices for the $k$ nearest neighbors of $\deformpoint^{0}_m$, and $\lle$ is a $\ndeform \times \ndeform$ adjacency matrix where $\lle_{ij}$ represent a edge between $\deformpoint^0_i$ and $\deformpoint^0_j$ with their corresponding linear weight if $j \in K_{i}$ and $0$ otherwise.  We then define a regularization term that penalize the deviation from the original local linear relationship:
\begin{equation}
\begin{split}
    E_{LLE}(\cpdw) &= \sumoverdeform \norm{\deformgmm_\gmmidx - \sum_{i \in K_\gmmidx}\lle_{\gmmidx i} \deformgmm_i}^2 
    \label{eqn:Elle}
\end{split}
\end{equation}
where $\deformgmm_i$ are the points from the CPD transformation (Eq.~\ref{eq:cpd_transform}).

\subsubsection{Geometric Prediction-based Regularization}
\label{sec:pred}

\begin{figure}[t]
    \centering
    \includegraphics[width=0.95\columnwidth]{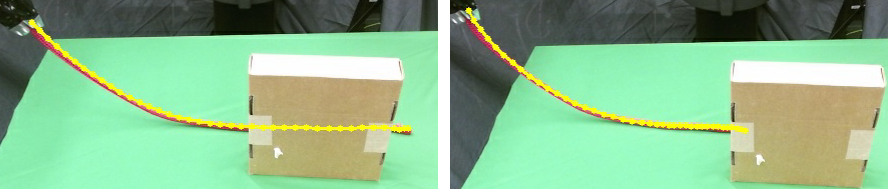}
    \caption{CPCPD producing a large change in the state when the end of the rope becomes occluded. Left: before. Right: after.} 
    \label{fig:cdcpd_occlusion_fail}
\end{figure}

While CDCPD is able to handle some occlusion cases, others like those shown in Fig.~\ref{fig:cdcpd_occlusion_fail} cause the algorithm to lose track of one side of the deformable object, shrinking the tracking to only the visible points. Though these two frames are close in time, the tracking result changes dramatically when the trailing end of the rope is occluded. Without any visual points to ``pin'' one side of the rope on the left side of the box, the CDCPD tracking moves all the points of the rope to the right side, effectively shrinking the object to fit it to the visible points.

To address this problem, we introduce a novel geometric prediction-based regularization term into the GMM-EM process. By using a geometric prediction like~\cite{robot:Dmitry, robot:Dale, robot:Mengyao}, we are able to avoid the downsides of using a physics simulation while capturing much of the behavior of a deformable object. Sec.~\ref{sec:pred_models} briefly details the models that we use in our results, but our method is agnostic to the specifics of the model used.

Let $\Deformpred$ be a prediction of the configuration of the object made by some model of deformable object motion. We define an additional cost function
\begin{equation}
\begin{split}
    E_{PRED}(\cpdw) 
    =& \sumoverdeform \norm*{\deformgmm_\gmmidx - \deformpred_\gmmidx}^2 \\
    =& \Tr \bigl((\Deformgmm - \Deformpred)^T(\Deformgmm - \Deformpred)\bigr) \\
    =& \Tr \bigl( (\Deformprev + \kernel \cpdw)^T(\Deformprev + \kernel \cpdw) \\
     &- 2 (\Deformprev + \kernel \cpdw)^T\Deformpred + {\Deformpred}^T\Deformpred \bigr)
    \label{eq:newocst}
\end{split}
\end{equation}
to encode the information from the model. Below we show how to incorporate Eq. \ref{eq:newocst} in the CDCPD regularization procedure.

Thus, similar to a tracking method like the Kalman filter, we make a prediction with the motion model, and then use it to update our tracking result along with what we observe at the next frame.  This approach has the advantage of not being too sensitive to large perceptual changes (e.g. the end of the rope disappearing), because the motion model helps to retain physical plausibility.

\subsubsection{Solving for the Transform Weights}

Adding all three regularization terms, we get the final cost function for the M-step:
\begin{equation}
\begin{split}
    Q&(\cpdw, \sigma^2) = \\
     &\sumovermasked \sumoverdeform \probgmmgivenmasked \frac{\norm*{\maskedpoint_\maskedidx - (\deformprev_\gmmidx + \kernel(\gmmidx,\cdot)\cpdw}^2}{2\sigma^2} \\
     & + \frac{3 N_{\prob}}{2}\log(\sigma^2) + \frac{\alpha}{2}E_{CPD}(\cpdw)\\
     & + \frac{\gamma}{2}E_{LLE}(\cpdw) + \frac{\zeta}{2}E_{PRED}(\cpdw),
    \label{cost}
\end{split}
\end{equation}
\noindent where $\alpha$, $\gamma$, and $\zeta$ are constant weights and $N_\prob=\sumovermaskeddeform \probgmmgivenmasked$.

Similar to the process shown in \cite{cv:GLTP}, we can minimize $Q$ in the M-step by computing a $\cpdw$ where $\partial Q/\partial W = 0$: Denote $\ones$ as a column vector of ones, and $d(v)$ as the diagonal form of vector $v$. 
Let $H = (\eye-\lle)^T(\eye-\lle)$. Then, 
\begin{equation}
\begin{split}
    A&=\left( d(\posteriorprob\ones)\kernel+\alpha\sigma^2 I+\gamma\sigma^2 H \kernel + \zeta \kernel \right) \\
    B&=\posteriorprob\Maskedpoints - (d(\posteriorprob\ones)+\gamma\sigma^2 H)\Deformgmm + \zeta (\Deformpred - \Deformgmm),
\end{split}
\end{equation}
\noindent where $A$ is a $M\times M$ matrix and $B$ is a $M\times 3$ matrix. Both $A$ and $B$ are calculated directly from terms we have already computed. We then obtain $\cpdw$ by solving
\begin{equation}
    \label{eq:WSolver}
    A \cpdw = B.
\end{equation}
\noindent Then we can obtain $\sigma^2$ using Eq.~(19) from \cite{robot:CDCPD}:
\begin{equation}
\begin{split}
    \label{eq:sigmaSolver}
    \sigma^2 = 
    & \frac{1}{3 N_{\prob}}\left( \Tr \left( \Maskedpoints^T d(\posteriorprob^T\ones)\Maskedpoints \right)\right) - 2\Tr \left( (\Deformgmm)^T\posteriorprob\Maskedpoints \right) \\
    & - 2 \Tr \left( \cpdw^T\kernel^T \posteriorprob \Maskedpoints \right) + \Tr \left( (\Deformgmm)^T d(\posteriorprob\ones) \Deformgmm \right) \\
    & + 2 \Tr \left( \cpdw^T\kernel^T d(\posteriorprob\ones) \Deformgmm\right) + \Tr \left( \cpdw^T \kernel^T d(\posteriorprob\ones) \kernel\cpdw \right)
\end{split}
\end{equation}
We repeat the E-step and M-step until it reaches the maximum number of iterations (100) or the change in $\sigma^2$ between two iterations is smaller than a threshold ($10^{-4}$). $\Deformgmm$ after one time step is thus $\Deformprev + \kernel\cpdw$.


\vspace{-0.05in} 
\subsection{Posterior Constraints}
\vspace{-0.05in} 

GMM-EM is able to reason about probabilities and costs, but the result is not guaranteed to be \textit{geometrically consistent}. For example, points could be further apart than is possible for a given deformable object. To address this, CDCPD introduced a post-processing optimization step which enforces some constraints. We start with those constraints (Sec.~\ref{sec:stretch}) and propose novel convex constraints for preventing self-intersection (Sec.~\ref{sec:self_inter}) and obstacle penetration (Sec.~\ref{sec:obstacle}). We emphasize that using convex constraints is essential for efficient and reliable optimization.

Denote $\deformpoint_\gmmidx^\text{GMM}$ as the $\gmmidx$th point of the GMM-EM result $\Deformgmm = \Deformprev + \kernel \cpdw$. Then our goal is to adjust every $\deformpoint_\gmmidx^\text{GMM}$ such that all points comply with the geometric constraints described above. 

\subsubsection{Known Correspondences and Stretching Limits}
\label{sec:stretch}

CDCPD considers two constraints. First, a constraint based on the maximum distance between the nodes in the mesh
\begin{equation}
    \norm*{\deformpoint_i - \deformpoint_j} \leq \lambda \geodistij \quad \forall (i, j) \in \Deformedges
\end{equation}
with a parameter $\lambda \geq 1$ which controls the flexibility of the constraint. Second, a constraint based on \textit{known correspondences}. For example, if we know that the grippers are rigidly connected to a particular part of the deformable object, then we can encode that constraint directly into the optimization at each timestep. We represent these correspondences between a set of known points $\{ z_1, \dots, z_K \} \subset \reals^{3K}$ and corresponding deformable point indices as $\correspondidx = [c_1, c_2, \dots, c_K] \in \naturals^{2K}$
\begin{equation}
    \deformpoint_m = z_k \quad \forall (m, k) \in \correspondidx \enspace .
\end{equation}

\subsubsection{Self-Intersection Constraints}
\label{sec:self_inter}

\begin{figure}[t]
    \centering
    \includegraphics[width=0.95\columnwidth]{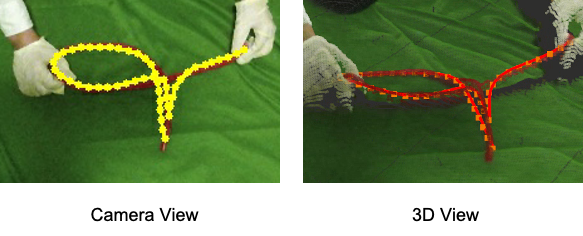}
    \caption{Example of CDCPD tracking result passing through itself when raised.}
    \label{fig:crossself}
\end{figure}

The first novel constraint we add is designed to prevent a tracking result from passing through itself (see Fig.~\ref{fig:crossself}). The general idea of preventing self intersection is checking pairs of edges for potential intersection and then constraining the movement of the points such that a small gap remains between any edges that could cross. This gap corresponds to the thickness of the object. Specifically, our algorithm checks all pairs of edges that do not share points. For two edges $\deformedge_i = (i_0,  i_1), \deformedge_j = (j_0, j_1)\in \Deformedges$, we regard them as two line segments and calculate the shortest distance $s_{ij}$ between the two closest points, $\deformnear{i}$ and $\deformnear{j}$, on these line segments. These points are:
\begin{align}
    \deformprevnear{i} &= r_i\Deformprev(i_0) + (1-r_i)\Deformprev(i_1) \\
    \deformprevnear{j} &= r_j\Deformprev(j_0) + (1-r_j)\Deformprev(j_1) \enspace ,
\end{align}
where $r_i$ and $r_j$ are two real number between 0 and 1. If $s_{ij}$ is smaller than the threshold $s_{check}$ we will constrain $\Deformpoints(i_0)$, $\Deformpoints(i_1)$, $\Deformpoints(j_0)$ and $\Deformpoints(j_1)$. The larger $s_{check}$ is, the less likely it is that we miss a self-intersection between two frames. However, a large  $s_{check}$ will likely result in more constraints in the optimization problem, which leads to a higher computation cost and potentially an infeasible optimization problem.

The closest points $\deformprevnear{i}$ and $\deformprevnear{j}$ found in $\Deformprev$ above correspond to the closest points $\deformnear{i}$ and $\deformnear{j}$ found in $\Deformpoints$. $\deformnear{i}$ and $\deformnear{j}$ are calculated as below:
\begin{align}
    \deformnear{i} &= r_i\Deformpoints(i_0) + (1-r_i)\Deformpoints(i_1) \\
    \deformnear{j} &= r_j\Deformpoints(j_0) + (1-r_j)\Deformpoints(j_1)
\end{align}
We constrain $\Deformpoints$ by requiring the projection of vector $\deformnear{i} - \deformnear{j}$ on the direction of $\deformprevnear{i} - \deformprevnear{j}$ to be larger than a collision threshold $s$. We write the constraints as below:
\begin{align}
    \label{eq:self_inter}
    (\deformnear{i}-\deformnear{j})^T \frac{\deformprevnear{i}-\deformprevnear{j}}{\norm*{\deformprevnear{i}-\deformprevnear{j}}} > s \quad \forall (i, j) \in \Deformedges
\end{align}
\noindent Intuitively, the parameter $s$ determines how far edges should keep away from each other, i.e. it approximates object thickness.

\subsubsection{Obstacle Interaction Constraints}
\label{sec:obstacle}

In addition to self-intersection constraints, we include a novel constraint to account for obstacle interaction. If we know the geometry of some obstacles in the scene, we can constrain the result of the tracking process to stay consistent with that known geometry. To achieve this, we add constraints based on the local geometry near each point $\deformprev_\gmmidx$. For each point $\deformprev_\gmmidx$, we find the nearest obstacle point $o_\gmmidx$ and the corresponding normal vector $n_\gmmidx$. We then constrain the tracking result to stay on the same side of the tangent plane defined by $o_\gmmidx$ and $n_\gmmidx$: $(\deformpoint_\gmmidx - o_\gmmidx)^T n_\gmmidx > 0 \quad \forall \deformpoint_\gmmidx \in \Deformpoints \enspace$.

Combining all the constraints together yields the following convex optimization problem:
\begin{equation}
\begin{split}
\label{eq:constrain}
    \argmin_{\Deformpoints} & \quad \sumoverdeform \norm*{\deformpoint_\gmmidx - \deformpoint_\gmmidx^\text{GMM}}^2\\
    \subjectto              & \quad \norm*{\deformpoint_i - \deformpoint_j} \leq \lambda \geodistij \quad \forall (i, j) \in \Deformedges \\
                            & \quad \deformpoint_m = z_k \quad \forall (m, k) \in \correspondidx  \\
                            & \quad (\deformnear{i}-\deformnear{j})^T  \frac{\deformprevnear{i}-\deformprevnear{j}}{\norm*{\deformprevnear{i}-\deformprevnear{j}}} > s \quad \\
                            & \quad (\deformpoint_\gmmidx-o_\gmmidx)^T n_\gmmidx > 0 \quad \forall \gmmidx \in [1, \dots, \ndeform] \enspace,
\end{split}
\end{equation}
which we solve with the  Gurobi~\cite{software:gurobi} optimization package, yielding our final estimate of $\Deformpoints$.


\begin{figure*}[t]
    \centering
    \includegraphics[width=\textwidth]{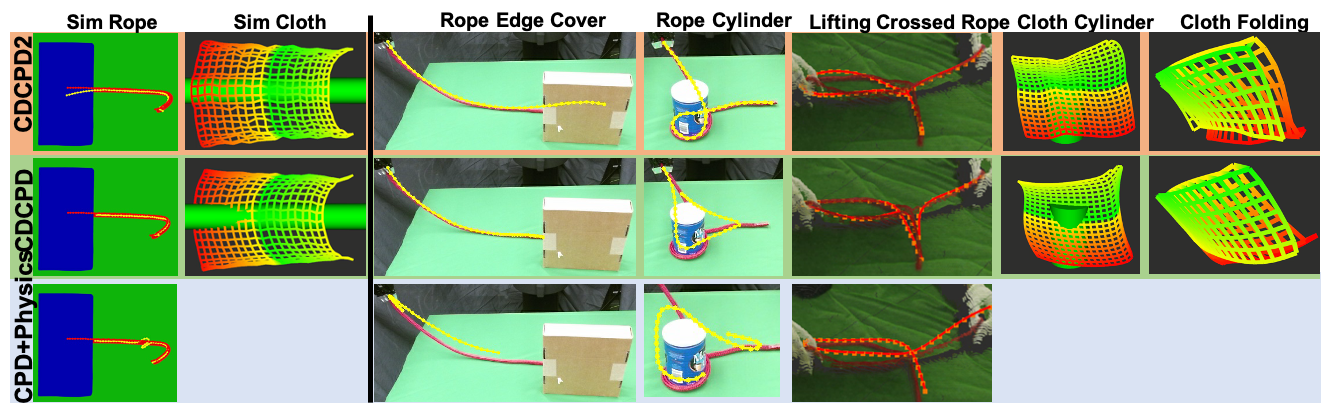}
    \caption{Comparison of results of \name{} and baselines for simulation (left of line) and real (right of line) data. The CPD+Physics implementation from the authors of \cite{robot:TangState} does not track cloth, so it is used only for rope. Columns show results for a single frame in each task. First column: ground truth (red) vs. estimates (yellow). \name{} used the diminishing rigidity model for all tasks shown except lifting crossed rope, where we used the ``no motion'' model because the rope was manipulated by a human (so $\gvel$ is unknown).}
    \label{fig:resultsfig}
    \vspace{-0.2in}
\end{figure*}

\vspace{-0.05in} 
\section{Experiments}
\vspace{-0.05in} 

We conducted experiments in simulation and the real world to analyze our algorithm quantitatively and qualitatively in the presence of severe occlusion and obstacles (see the accompanying video). 
We compare our results with CPD+physics \cite{robot:TangState} and CDCPD \cite{robot:CDCPD}.

The parameter values are: $\beta =1.0$, $\alpha= 0.5$, $\gamma=1.0$
, $\zeta= 2.0$, $k_{vis}=100$ (see Eq. (7) of \cite{robot:CDCPD}), $s_{check} = 0.02m$, $s= 0.01m$, and $\lambda=1.1$. Point clouds are downsampled using a voxel grid filter with a grid size of 2cm. \name{}, CDCPD and CPD+Physics are implemented in C++ and tested on an Intel i7-8700 @ 3.7GHz processor with 32 GB RAM.

\vspace{-0.05in} 
\subsection{Geometric Prediction Models}
\vspace{-0.05in} 
\label{sec:pred_models}

Because our method relies on a prediction of deformable object motion, we need an efficient model that outputs this kind of prediction. We emphasize that the model need not be very accurate, but only that it is more physically plausible than the output of the CPD process alone. To show our algorithm's performance with (and robustness to) a variety of models, we provide results for three different geometric models of motion. First, the most naive model, we call \textit{No Motion}, assumes there is no movement between frames: $\Deformpred = \Deformprev$. While naive, this model may be reasonable when the movement between two frames is small. The second prediction model we evaluate is the \textit{diminishing rigidity  model}~\cite{robot:Dmitry, robot:Dale}: $
    \Deformpred = \Deformprev + J(\Deformpoints, \gconfig) \gvel$, where $J(\Deformpoints, \gconfig)$ is an estimate of the Jacobian mapping gripper movements to deformable object movement. We use $k=10.0$ (see \cite{robot:Dmitry}). The third model prediction we tried is a \textit{constrained directional rigidity model}~\cite{robot:Mengyao}. It builds on diminishing rigidity by directly accounting for the direction of gripper motion, and formulating constraints to account for obstacles. This results in a more expressive Jacobian: $    \Deformpred = \Deformprev + J(\Deformpoints, \gconfig, \gvel, \obstacles) \gvel$. We use $k_r=10.0$, $k_g=5.0$ and $k_D=5.0$ (see \cite{robot:Mengyao}).

\subsection{Experiments with Simulated Data}
To analyze performance quantitatively, we obtained ground truth from simulation in Blender and compared all methods to this. The rope is modelled as a soft body with 49 line segments. The cloth is modelled as a 20 by 20 mesh. RGBD images are rendered using the Eevee rendering engine with size 810 $\times$ 540.

\begin{figure}[t]
    \centering
    \includegraphics[width=\columnwidth]{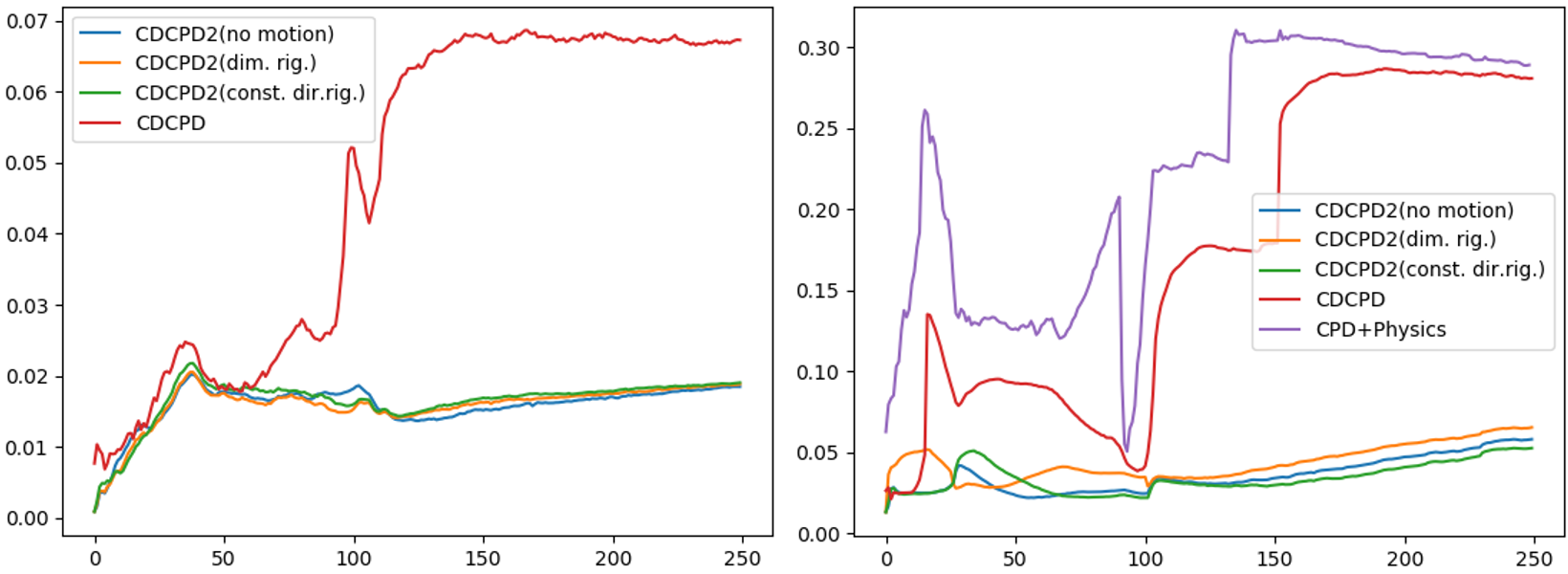}
    \caption{Mean distance error (m) vs. frame index for simulation experiments. Left: cloth covering cylinder. Right: rope edge-covering.}
    \label{fig:baseline}
\end{figure}

The first experiment demonstrates the ability of our method to prevent the tracking result from shrinking to the visible part of the object (as in Fig. \ref{fig:cdcpd_occlusion_fail}). We drag one end of the rope until the free end of the rope is occluded by the obstacle. We can see from Fig. \ref{fig:resultsfig} that our result won't shrink (regardless of the motion model used), while CDCPD and CPD+Physics will shrink quickly. Fig. \ref{fig:baseline} shows the comparison with ground truth. We can see when the free end is not occluded, i.e. the first several frames, the error of all these methods is close. However, when the free end is occluded, the error of CDCPD and CPD+Physics becomes much larger.


Our second experiment demonstrates our algorithm's ability to handle interaction with obstacles, Here we drape a cloth over a cylinder with a camera looking from above. Our result doesn't penetrate the cylinder, while CDCPD's does, as shown in Fig. \ref{fig:resultsfig}.




\vspace{-0.05in} 
\subsection{Experiments with real data}
\vspace{-0.05in} 

We performed several experiments with real data to guage the qualitative performance of our method. Examples are shown in Fig. \ref{fig:resultsfig} and the accompanying video. The experiments showed manipulation of a rope and cloth under occlusion and interacting with obstacles. Due to space limitations we do not describe each experiment in detail but instead present the key observations: 1) For both rope and cloth, our method was robust to edge-covering (i.e. not shrinking when an edge was covered), unlike CDCPD and CPD+Physics; 2) \name{} was effective in keeping the object out of known obstacles (e.g. Fig. \ref{fig:intro}) while CDCPD was not; 3) \name{} prevented edge crossings for rope on-par with CPD+Physics and better than CDCPD; 4) For many experiments, the choice of motion model did not have a major impact, however the best-performing model overall was diminishing rigidity. 

\vspace{-0.05in} 
\subsection{Computation time}
\vspace{-0.05in} 

The average computation time per frame of \name{} across all tasks is 26 ms (rope) and 193 ms (cloth). For CDCPD, the average computation time is 22 ms (rope) and 101 ms (cloth). For CPD+Physics, the average computation time is 23 ms (rope). The increase in computation time for our method on cloth is largely due to the additional obstacle interaction and self-intersection constraints, which take time to compute and entail a more difficult optimization problem.

\vspace{-0.05in} 
\section{Conclusion}
\vspace{-0.05in} 

Our results show the ability of our algorithm to handle obstacle interaction, self-intersection and severe occlusion when tracking deformable objects better than previous work. We increased the robustness to occlusion by introducing a prediction-based regularization term in GMM-EM, which is inspired by the Kalman filter. By adding posterior constraints to the result from GMM-EM, we prevented the result from penetrating obstacles and intersecting itself. In future work, we may explore learning a good initialization of $\cpdw$ by training on a dataset collected in a simulation environment.


\vspace{-0.05in} 
\bibliographystyle{IEEEtran}
\bibliography{references_strings_abbrev,references}

\end{document}